\documentclass[11pt,a4paper]{article}
\usepackage[hyperref]{ranlp2023}
\usepackage{times}
\usepackage{latexsym}

\usepackage{microtype}

\usepackage{multirow}
\usepackage[nolist]{acronym}
\usepackage{subcaption}
\usepackage{graphicx}
\usepackage[mathscr]{euscript}
\DeclareSymbolFont{rsfs}{U}{rsfs}{m}{n}
\DeclareSymbolFontAlphabet{\mathscrsfs}{rsfs}
\usepackage{algorithm}
\usepackage[noend]{algpseudocode}
\usepackage{adjustbox,booktabs}
\usepackage{dblfloatfix} 

\aclfinalcopy 

\title{AspectCSE: Sentence Embeddings for Aspect-based Semantic Textual Similarity Using Contrastive Learning and Structured Knowledge}

\author{Tim Schopf$^1$, Emanuel Gerber$^1$, Malte Ostendorff$^2$, and Florian Matthes$^1$ \\
         $^1$Technical University of Munich, Department of Computer Science, Garching, Germany \\
         $^2$DFKI GmbH, Berlin, Germany \\ 
         \texttt{\{tim.schopf,emanuel.gerber,matthes\}@tum.de} \\
         \texttt{malte.ostendorff@dfki.de}}

\date{}

\begin{acronym}
\acro{nlp}[NLP]{natural language processing}
\acro{lm}[LM]{Language Model}
\acro{plm}[PLM]{Pretrained Language Model}
\acroplural{plm}[PLMs]{Pretrained Language Models}
\acro{sts}[STS]{semantic textual similarity}
\acro{kg}[KG]{knowledge graph}
\acroplural{kg}[KGs]{knowledge graphs}
\acro{pwc}[PwC]{Papers with Code}
\acro{p}[P]{Precision}
\acro{r}[R]{Recall}
\acro{mrr}[MRR]{Mean Reciprocal Rank}
\end{acronym}

\begin{document}
\maketitle
\begin{abstract}
Generic sentence embeddings provide a coarse-grained approximation of semantic textual similarity but ignore specific aspects that make texts similar. Conversely, aspect-based sentence embeddings provide similarities between texts based on certain predefined aspects. Thus, similarity predictions of texts are more targeted to specific requirements and more easily explainable. In this paper, we present AspectCSE, an approach for aspect-based contrastive learning of sentence embeddings. Results indicate that AspectCSE achieves an average improvement of 3.97\% on information retrieval tasks across multiple aspects compared to the previous best results. We also propose using Wikidata knowledge graph properties to train models of multi-aspect sentence embeddings in which multiple specific aspects are simultaneously considered during similarity predictions. We demonstrate that multi-aspect embeddings outperform single-aspect embeddings on aspect-specific information retrieval tasks. Finally, we examine the aspect-based sentence embedding space and demonstrate that embeddings of semantically similar aspect labels are often close, even without explicit similarity training between different aspect labels.
\end{abstract}

\section{Introduction}

\begin{figure*}[ht!]
    \centering
    \begin{subfigure}{0.32\textwidth}
        \includegraphics[width=\textwidth]{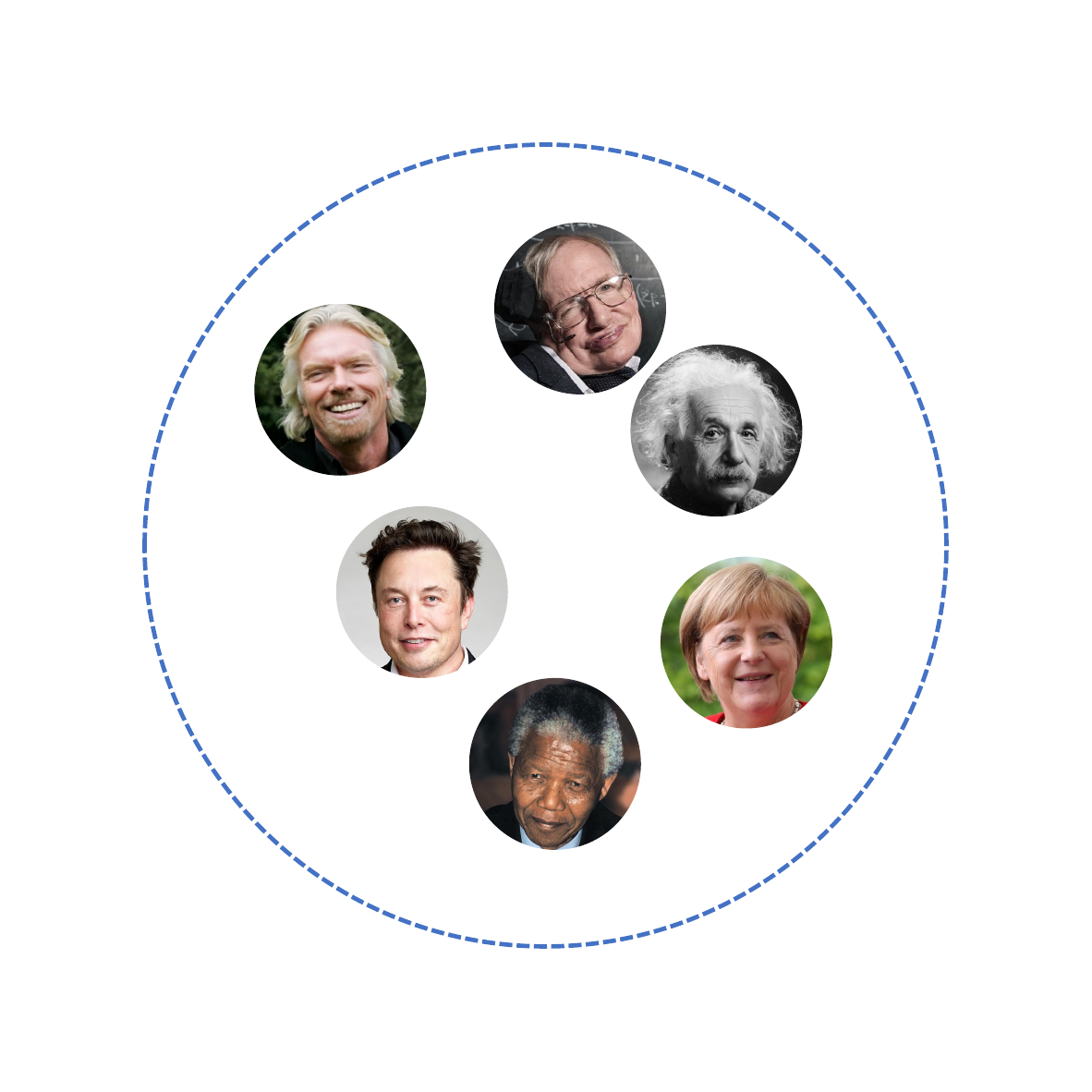}
        \caption{Generic sentence embeddings}
        \label{fig:first}
    \end{subfigure}
    \hfill
    \begin{subfigure}{0.32\textwidth}
        \includegraphics[width=\textwidth]{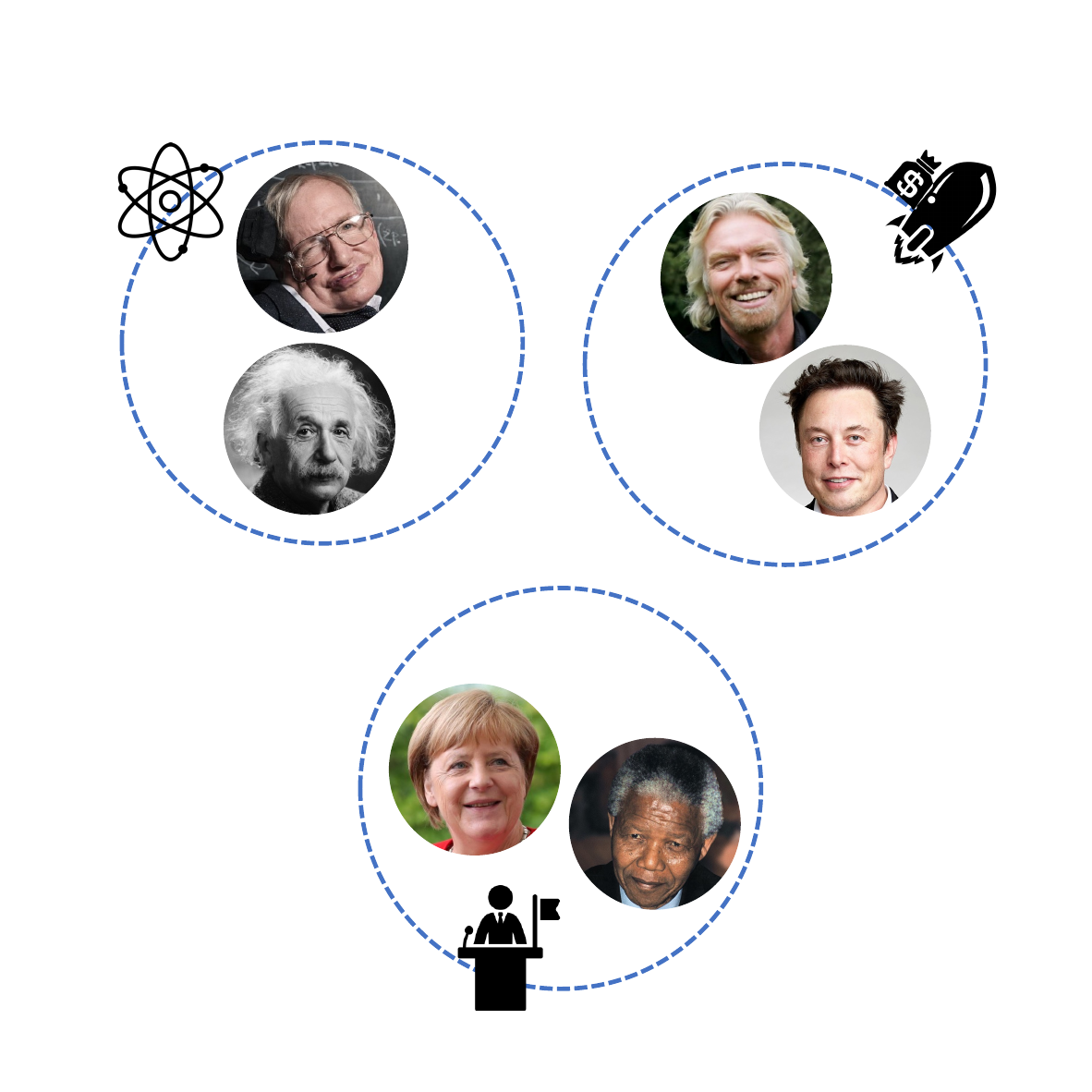}
        \caption{Sentence embeddings based on the \textit{profession} aspect.}
        \label{fig:second}
    \end{subfigure}
    \hfill
    \begin{subfigure}{0.32\textwidth}
        \includegraphics[width=\textwidth]{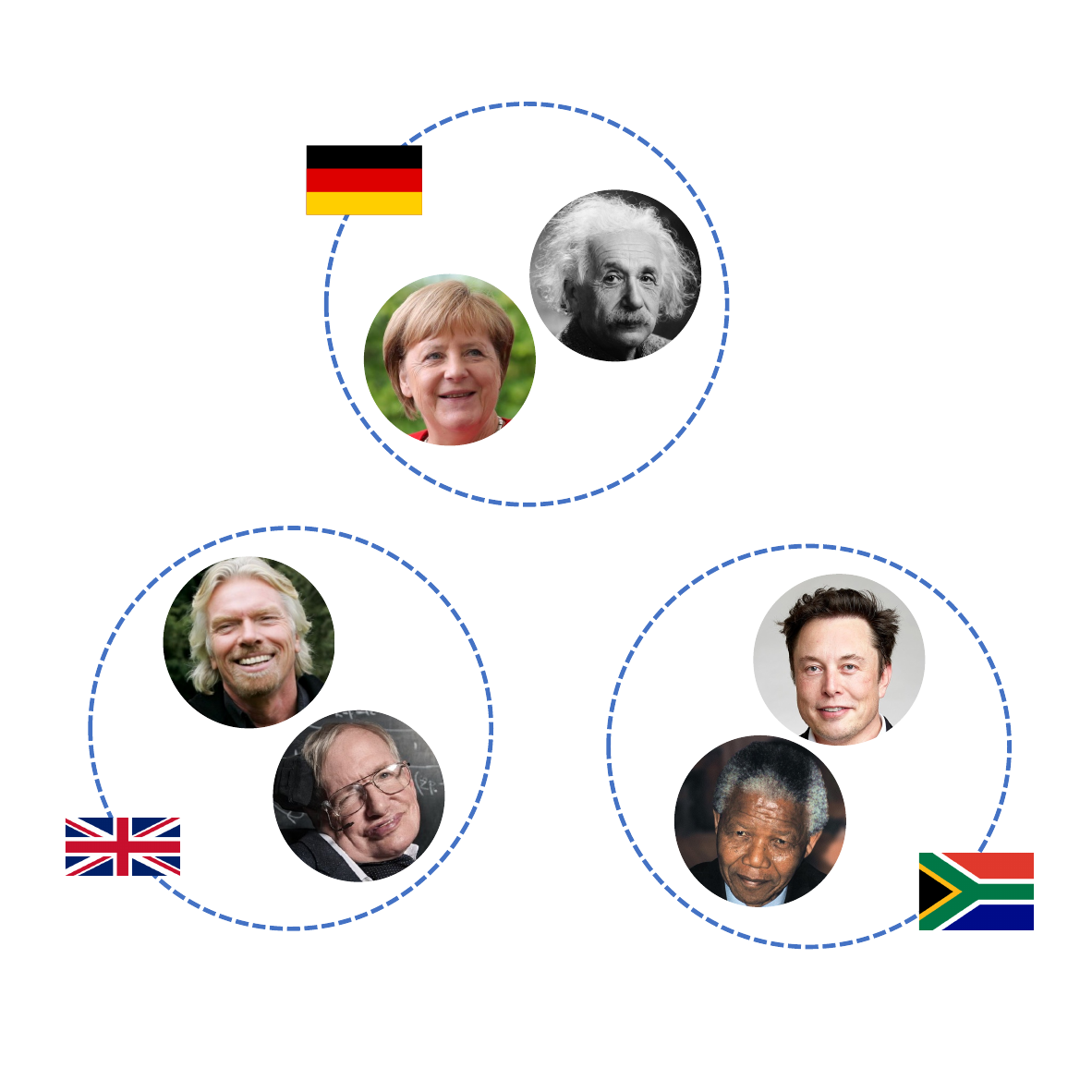}
        \caption{Sentence embeddings based on the \textit{country of birth} aspect.}
        \label{fig:third}
    \end{subfigure}
    \caption{Images of famous people with the corresponding Wikipedia introductory texts as sentence embeddings in a dense vector space. Blue dashed circles represent clusters of semantically similar embeddings. Based on the encoded aspect, embeddings of these same texts can be distributed differently in a vector space. (a) All generic embeddings are close and approximately evenly distributed as the texts introduce famous people. (b) Embeddings that focus on the \textit{profession} aspect are close if the people have similar professions. (c) Embeddings that focus on the \textit{country of birth} aspect are close if the people have similar countries of birth.}
    \label{fig:aspect-based-similarity-example}
\end{figure*}

Sentence embeddings are representations of sentences or short text paragraphs in a dense vector space, such that similar sentences are close to each other \cite{reimers-gurevych-2020-making}. Learning sentence embeddings is a fundamental task in \ac{nlp} and has already been extensively investigated in the literature \cite{NIPS2015_f442d33f,hill-etal-2016-learning,conneau-etal-2017-supervised,logeswaran2018an,cer-etal-2018-universal,reimers-gurevych-2019-sentence,gao-etal-2021-simcse,schopf2023efficient}. Generic sentence embeddings can be used to distinguish between similar and dissimilar sentences, without considering which aspects of sentences are similar \cite{ostendorff-etal-2020-aspect}. Moreover, they are often evaluated on generic \ac{sts} tasks \cite{marelli-etal-2014-sick,agirre-etal-2012-semeval,agirre-etal-2013-sem,agirre-etal-2014-semeval,agirre-etal-2015-semeval,agirre-etal-2016-semeval,cer-etal-2017-semeval} in which sentence similarity scores rely on human annotations. However, the concept of generic \ac{sts} is not well defined, and text similarity depends heavily on the aspects that make them similar \cite{bar-etal-2011-reflective,10.1145/3383583.3398525,10.1145/3529372.3530912}. We follow the argument of \citet{bar-etal-2011-reflective} on textual similarity and define \textit{aspects} as inherent properties of texts that must be considered when predicting their semantic similarity. Based on the different aspects focused on in texts, their similarities can be perceived very differently. Figure \ref{fig:aspect-based-similarity-example} illustrates an example of aspect-based \ac{sts}. For example, Wikipedia introduction texts of famous individuals can generally be considered similar as all texts introduce people who are known to the public. However, focusing the comparison on specific aspects (e.g., \textit{country of birth} or \textit{profession}) leads to different semantic similarity assessments for the same texts. Although Wikipedia is a special case as the introduction texts represent specific entities, this characteristic can nevertheless be generalized to different aspects found in any text. When deciding the similarity of texts, different aspects must be considered. Consequently, human-annotated \ac{sts} datasets introduce considerable subjectivity regarding the evaluated aspects.

Prior work uses siamese networks and a multiple negative ranking loss \cite{https://doi.org/10.48550/arxiv.1705.00652} with only positive samples from the train set to create sentence embeddings for single aspects \cite{10.1145/3529372.3530912}. Sentence embeddings for single aspects only consider one specific aspect during similarity comparisons. Using structured knowledge from \acp{kg} for language model training has been shown to improve performances on all types of downstream tasks \cite{schneider-etal-2022-decade} and also provides the possibility to create sentence embeddings that focus on multiple specific aspects simultaneously. These sentence embeddings are especially useful in information retrieval or unsupervised text classification settings \cite{schopf_etal_webist_21,schopf_etal_kdir_22,schopf2023exploring,10.1145/3582768.3582795,10.1007/978-3-031-24197-0_4}.

In this work, we advance state-of-the-art sentence embeddings for aspect-based \ac{sts} using AspectCSE, an approach for \underline{aspect}-based \underline{c}ontrastive learning of \underline{s}entence \underline{e}mbeddings. Additionally, we introduce multi-aspect sentence embeddings that simultaneously consider multiple specific aspects during similarity comparisons. We show the effectiveness of multi-aspect sentence embeddings for both information retrieval and exploratory search tasks. Finally, we demonstrate that using \ac{kg} properties can be extremely beneficial for creating both single- and multi-aspect sentence embeddings. 

\begin{figure*}[ht!]
    \centering
    \includegraphics[width=\textwidth]{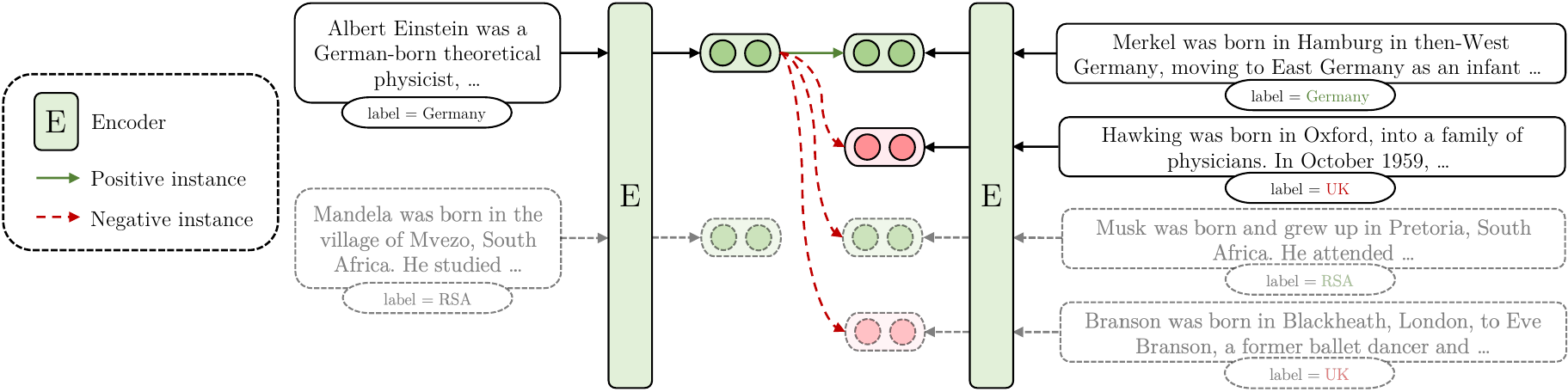}
    \caption{AspectCSE uses (\textit{anchor, positive, negative}) triplets to train aspect-specific sentence embedding models. Pairs with the same label for a specific aspect (here: \textit{country of birth}) are used as positives and those with different labels for the same aspect and other in-batch instances as negatives.}
    \label{fig:aspect-based-contrastive-learning}
\end{figure*}

\section{Related Work}

In \ac{nlp}, \textit{aspects} are most commonly examined in sentiment analysis problems \cite{pontiki-etal-2014-semeval,xue-li-2018-aspect,brun-nikoulina-2018-aspect,zhang-etal-2021-towards-generative,yan-etal-2021-unified,liang-etal-2022-bisyn}. Thus, the goal is to identify the aspects of given target entities and the sentiment expressed for each aspect \cite{pontiki-etal-2014-semeval}. 

Some works investigate aspect-based \ac{sts} by considering it as a segmentation task. \citet{10.1145/3274300} first segmented abstracts of research papers according to different aspects. Then, they constructed semantic representations from these aspect-based segments, which can be used to find analogies between research papers. \citet{huang-etal-2020-coda} presented a human-annotated dataset that segments 10,966 English abstracts in the COVID-19 Open Research Dataset \cite{wang-etal-2020-cord} by the aspects background, purpose, method, result/contribution, and others. \citet{10.1145/3197026.3197059} learned multi-vector representations of segmented scientific articles in which each vector encodes a different aspect. However, segmenting texts can harm their coherence and decrease the performance of downstream \ac{nlp} models \cite{gong-etal-2020-recurrent}.

Other approaches propose to treat aspect-based \ac{sts} as a pairwise multi-class classification problem \cite{ostendorff-etal-2020-aspect,10.1145/3383583.3398525}. However, \citet{reimers-gurevych-2019-sentence} argue that pairwise classification with transformer models results in quadratic complexity. Therefore, this approach is not suitable for large-scale \ac{sts} tasks.   

To address the issues using previous approaches, \citet{10.1145/3529372.3530912} proposed training aspect-based embeddings for research papers. In this work, we use AspectCSE and \ac{kg} properties to train single- and multi-aspect sentence embeddings. This allows us to focus on multiple specific aspects simultaneously while improving the performance of aspect-based sentence embeddings in \ac{sts} tasks.

\section{Embedding Methods}

\subsection{AspectCSE} \label{sec:contrastive-learning}

Recently, contrastive learning has exhibited state-of-the-art performance for generic sentence embeddings \cite{gao-etal-2021-simcse,giorgi-etal-2021-declutr,chuang-etal-2022-diffcse}. The contrastive learning objective creates effective representations by pulling semantically close neighbors together and pushing apart non-neighbors \cite{1640964}. We follow the proposed supervised contrastive learning framework of \citet{gao-etal-2021-simcse} and use a cross-entropy-loss with negatives per anchor-positive pair and random in-batch negatives. To train aspect-based sentence embedding models, we assume a set of triplets $\mathcal{D} = \{(x_{i}^{a},x_{i}^{a+},x_{i}^{a-})\}$. Here, $x_{i}^{a}$ is an anchor sentence, $x_{i}^{a+}$ is semantically related, and $x_{i}^{a-}$ is semantically unrelated to $x_{i}^{a}$ with respect to aspect $a$. With $\textbf{h}_{i}^{a}$, $\textbf{h}_{i}^{a+}$, and $\textbf{h}_{i}^{a-}$ as representations of $x_{i}^{a}$,$x_{i}^{a+}$, and $x_{i}^{a-}$, the training objective with a mini-batch of $N$ triplets is expressed as:

\begin{equation}\label{eq:contrastive-loss}
\resizebox{.89\linewidth}{!}{$
  \ell_{i} = -\log \frac{e^{sim(\textbf{h}_{i}^{a},\textbf{h}_{i}^{a+})/\tau}}{\sum_{j=1}^{N}(e^{sim(\textbf{h}_{i}^{a},\textbf{h}_{j}^{a+})/\tau} + e^{sim(\textbf{h}_{i}^{a},\textbf{h}_{j}^{a-})/\tau})}$}
\end{equation}

where $\tau$ is a temperature hyperparameter and $sim(\textbf{h}_1, \textbf{h}_2)$ is the cosine similarity $\frac{\textbf{h}_1 \cdot \textbf{h}_2}{|| \textbf{h}_1|| \cdot || \textbf{h}_2||}$. To encode input sentences, we use BERT-based pretrained language models \cite{devlin-etal-2019-bert} and fine-tune the parameters using the contrastive objective (Equation \ref{eq:contrastive-loss}). Figure \ref{fig:aspect-based-contrastive-learning} illustrates the proposed AspectCSE approach.

\subsection{Multiple Negative Ranking Using Anchor-Positive Pairs Only} \label{sec:multiple-negative-ranking}

As a baseline, we perform aspect-based fine-tuning of BERT-based pretrained language models following the state-of-the-art approach of \citet{10.1145/3529372.3530912}. Therefore, we use mean pooling and a multiple negative ranking loss \cite{https://doi.org/10.48550/arxiv.1705.00652} with anchor-positive pairs for training. Therefore, the training input comprises a set of positive samples $\mathcal{D} = \{(x_{i}^{a},x_{i}^{a+})\}$ only. During training, every instance $x_{j}^{a+} = \{x_{1}^{a+}... x_{N-1}^{a+}\}$ within a mini-batch of $N$ samples is used as random negative for anchor $x_{i}^{a}$ if $i \neq j$.

\section{Data}

For our experiments, we use two different datasets. First, we use a benchmark dataset derived from \ac{pwc} \footnote{\href{https://paperswithcode.com}{https://paperswithcode.com}} to evaluate the effectiveness of AspectCSE. We also use Wikipedia and the Wikidata \ac{kg} \cite{10.1145/2629489} to build a dataset for learning multi-aspect sentence embeddings. In all our experiments, we consider a pair of texts as positive if they share the same label for a particular aspect. Accordingly, negatives comprise a pair of texts with different labels for a particular aspect.

\subsection{Papers with Code} \label{sec:pwc-section}

The \ac{pwc} dataset is a collection of research paper abstracts that are annotated with \textit{task}, \textit{method} and \textit{dataset} aspects and their respective labels \cite{10.1145/3529372.3530912}. In this dataset, for example, a label of the \textit{task} aspect is \textit{self-supervised learning} or \textit{machine translation}. We obtain the dataset version from 2022-05-25 
and remove paper abstracts that belong to aspect labels with more than 100 instances. Abstracts with less than 100 characters are also removed. Table \ref{tab:pwc-data} summarizes the resulting \ac{pwc} dataset. We split the final \ac{pwc} dataset into 80\% training and 20\% test paper abstracts for our experiments.

\begin{table}[ht!]
    \centering
    \begin{tabular}{lrr}
    \hline
    \textbf{Aspect} & \textbf{\# Papers} & \textbf{\# Labels} \\ \hline
    Task            & 32,873             & 2,481              \\
    Method          & 10,213             & 1,724              \\
    Dataset         & 7,305              & 3,611              \\ \hline
    \end{tabular}
    \caption{Summary of the \ac{pwc} dataset.}
    \label{tab:pwc-data}
\end{table}

\subsection{Wikipedia and Wikidata} \label{sec:wikipedia-wikidata}

Wikipedia contains a broad range of topics with many possible aspects for each article. We have found that the number of articles regarding companies in Wikipedia accounts for a large portion of the articles, while the introductory sections contain a reasonable amount of different aspects. Therefore, in our experiments focus on a subset of Wikipedia, which includes the introduction section of articles about companies only. Furthermore, we use the commonly occurring aspects \textit{industry (e.g., What type of product/service does the company offer?)} and \textit{country (e.g., What country is the company based in?)} for our experiments. Since Wikipedia comprises unstructured texts only, we take advantage of most Wikidata \ac{kg} entities being linked to their corresponding Wikipedia articles. We also consider specific Wikidata properties as aspects while using the values linked to a seed article by the specific properties as labels. In this case, we use the Wikidata properties \textit{country} and \textit{industry} as aspects while taking the values linked to the company articles by these properties as labels. Therefore, we follow the approach in Algorithm \ref{alg:create-dataset} to construct our dataset.

\begin{algorithm}
    \caption{Construct aspect-based dataset}
    \label{alg:create-dataset}
    \begin{algorithmic}[0]
    \Require
    \State {$companies=$ list of all Wikidata entities $e$ of type \textit{business} (Q4830453)}
    \State {$companies_{annotated} \gets \emptyset$}
    \Procedure{annotate}{$companies$} 
        \For{$e$ \textbf{in} $companies$}
            \If{$e_{k}$ has Wikipedia article $w_{k}$} 
                \State {$s=$ introduction section of $w_{k}$}
                \State {$s_{c}=$ \textit{country} (P17) value(s) of $e_{k}$}
                \State {$s_{i}=$ \textit{industry} (P452) value(s) of $e_{k}$}
                \State {$companies_{annotated} \mathrel{+}= (s,s_{c},s_{i})$}
            \EndIf
        \EndFor
        \Return {$companies_{annotated}$}
    \EndProcedure
    \end{algorithmic}
\end{algorithm}

We use the Wikidata SPARQL API to find the companies as well as the country and industry values linked to them. We also use the Kensho Derived Wikimedia Dataset\footnote{\href{ https://www.kaggle.com/datasets/kenshoresearch/kensho-derived-wikimedia-data}{https://www.kaggle.com/datasets/kenshoresearch/kensho-derived-wikimedia-data}}, which comprises preprocessed Wikipedia and Wikidata dumps from 2019-12-01, to obtain the Wikipedia introduction sections of the retrieved companies. Moreover, we utilize the Kensho Derived Wikimedia Dataset to sample 10,000 random articles from different topics without any aspect information. In addition to the company introduction sections, these random articles are used as further negatives during training. This ensures that the model learns to distinguish between different aspect labels and between different topics. Table \ref{tab:wikidata-data} summarizes the resulting dataset. For example, the labels for the \textit{country} aspect are USA or Germany. For our experiments, we split the final dataset into 80\% training and 20\% test data.

\begin{table}[ht!]
    \centering
    \begin{tabular}{lrr}
    \hline
    \textbf{Aspect} & \textbf{\# Articles} & \textbf{\# Labels} \\ \hline
    Industry            & 6,082  &  97          \\
    Country          & 2,062  &  75          \\ 
    \textit{Random articles} & 10,000  &  -         \\\hline
    \end{tabular}
    \caption{Summary of the Wikipedia + Wikidata dataset.}
    \label{tab:wikidata-data}
\end{table}

\begin{table*}[ht!]
    \centering
    \resizebox{1.94\columnwidth}{!}{%
    \renewcommand{\arraystretch}{1.2} %
   
    \begin{adjustbox}{center}

    \begin{tabular}{ccl ccc ccc ccc}
    \toprule
    \multicolumn{3}{l}{\textbf{Aspects $\rightarrow$}} & \multicolumn{3}{c}{\textit{Task}} & \multicolumn{3}{c}{\textit{Method}} & \multicolumn{3}{c}{\textit{Dataset}} \\

    \cmidrule(lr){4-6}
    \cmidrule(lr){7-9}
    \cmidrule(lr){10-12}
    
    \multicolumn{3}{l}{\textbf{Methods $\downarrow$}} &      \textbf{P} &      \textbf{R} &    \textbf{MRR} &        \textbf{P} &      \textbf{R} &    \textbf{MRR} &          \textbf{P} &      \textbf{R} &    \textbf{MRR}  \\
    
    \midrule

    \multirow{3}{*}{\rotatebox[origin=c]{90}{\parbox[c]{1cm}{\centering \small{Generic \newline}}}} & \multicolumn{2}{l}{$\textrm{SciBERT}_{\small{\texttt{base}}}$} & 
    0.071 &  0.070 & 0.244 &  0.051 &  0.056 & 0.181 & 0.060 & 0.101 & 0.212 \\
    & \multicolumn{2}{l}{$\textrm{DeCLUTR}_{\small{\texttt{sci-base}}}$} & 
    0.130 &  0.131 & 0.369 &  0.069 &  0.078 & 0.219 & 0.099 & 0.170 & 0.317 \\
    & \multicolumn{2}{l}{SPECTER} & 
    0.248 &  0.247 & 0.521 &  0.104 &  0.117 & 0.277 & 0.183 & 0.311 & 0.464 \\
    \hline
    \multirow{2}{*}{\rotatebox[origin=c]{90}{\parbox[c]{1cm}{\centering \small{Aspect-based}}}}
     & \multicolumn{2}{l}{Multiple Negative Ranking} & 0.409 & 0.424 & 0.768 & 0.263 & 0.302 & 0.595 & 0.172 & 0.418 & 0.465 \\ 
    {} & \multicolumn{2}{l}{$*$ AspectCSE} & \textbf{0.416} & \textbf{0.431} & \textbf{0.776} & \textbf{0.268} & \textbf{0.312} & \textbf{0.606} & \textbf{0.186} & \textbf{0.461} & \textbf{0.507} \\ 
    \bottomrule
    \end{tabular}
\end{adjustbox}%
}
\caption{Evaluation results for retrieving the $k=10$ most similar elements for different sentence embedding approaches on the \ac{pwc} test dataset. \textit{AspectCSE} indicates the training approach explained in Section \ref{sec:contrastive-learning}. \textit{Multiple Negative Ranking} indicates the training approach explained in Section \ref{sec:multiple-negative-ranking}. Precision@k (P), Recall@k (R), and Mean Reciprocal Rank@k (MRR) are reported.} 
\label{tab:pwc-evaluation}
\end{table*}

To train aspect-based sentence embeddings with AspectCSE, we further process the dataset to yield triplets as follows:

\begin{itemize}
    \item \textbf{Single-aspect-specific (Country)}: \\ $(x_{i}^{a},x_{i}^{a+},x_{i}^{a-})$ $\Rightarrow$ $x_{i}^{a+}$ and $x_{i}^{a-}$ are positive and negative samples w.r.t. the country aspect $a$.
    \item \textbf{Single-aspect-specific (Industry)}: \\ $(x_{i}^{b},x_{i}^{b+},x_{i}^{b-})$ $\Rightarrow$ $x_{i}^{b+}$ and $x_{i}^{b-}$ are positive and negative samples w.r.t. the industry aspect $b$.
    \item \textbf{Multi-aspect-specific (Intersection)}: \\ $(x_{i}^{a,b}, x_{i}^{a+\cap{b+}}, x_{i}^{a-\cap{b-}})$ $\Rightarrow$ $x_{i}^{a+\cap{b+}}$ is a positive sample if it has \textbf{both} the same country aspect $a$ \textbf{and} the same industry aspect $b$ as the seed sentence.
    \item \textbf{Multi-aspect-specific (Union)}: \\ $(x_{i}^{a,b}, x_{i}^{a+\cup{b+}}, x_{i}^{a-\cup{b-}})$ $\Rightarrow$ $x_{i}^{a+\cup{b+}}$ is a positive sample if it has \textbf{either} the same country aspect $a$ \textbf{or} the same industry aspect $b$ as the seed sentence. 
\end{itemize}

\section{Experiments}


\subsection{Comparison with Baselines}

To evaluate AspectCSE against state-of-the-art baselines, we use the \ac{pwc} benchmark dataset described in Section \ref{sec:pwc-section} for model training and testing.

\paragraph{Generic Sentence Embeddings} We evaluate AspectCSE against multiple generic sentence embedding models from the scholarly domain. These models are pretrained on scientific literature and produce domain-specific state-of-the-art sentence embeddings without leveraging any aspect information. We use SciBERT \cite{beltagy-etal-2019-scibert}, SPECTER \cite{cohan-etal-2020-specter}, and DeCLUTR \cite{giorgi-etal-2021-declutr} in their $\texttt{base}$-versions as published by their authors without any fine-tuning on our corpus. For SciBERT, we use the concatenated outputs of the last four layers as embeddings.

\begin{table}[htbp]
    \renewcommand{\arraystretch}{0.75}
    \centering
    \begin{tabular}{ |p{4cm}|p{2cm}| }
    \hline
    \multicolumn{1}{|c}{\bfseries \small Parameter} & \multicolumn{1}{|c|}{\bfseries \small Value} \\
    \hline
    \small Training epochs & \small $3$ \\
    \hline
    \small Batch size & \small $14$ \\
    \hline
    \small Learning rate & \small $5e-5$ \\
    \hline
    \small Max sequence length & \small $320$ \\
    \hline
    \small Pooler type & \small CLS \\
    \hline
    \small Temperature for softmax & \small $0.05$ \\
    \hline
    \small Floating precision & \small $16$ \\
    \hline
    \end{tabular}
    \caption{AspectCSE fine-tuning configuration.}
    \label{tab:training_configuration_simcse}
\end{table}

\paragraph{Aspect-based Sentence Embeddings} In addition to generic baselines, we train aspect-based sentence embedding models for each \ac{pwc} aspect using SciBERT and the multiple negative ranking approach, as described in Section \ref{sec:multiple-negative-ranking}. To train AspectCSE, we use SciBERT as base model and the fine-tuning configuration presented in Table \ref{tab:training_configuration_simcse}. For aspect-specific baseline training with multiple negative ranking, we use the same configuration, except that we follow the approach of \citet{10.1145/3529372.3530912}, and apply MEAN pooling. For AspectCSE, we follow the argument of \citet{gao-etal-2021-simcse}, who found that different pooling methods do not matter much and use CLS.

\subsection{Multi-aspect Sentence Embeddings}\label{sec:multi-aspect-sentence-embeddings}

We use the Wikipedia + Wikidata dataset described in Section  \ref{sec:wikipedia-wikidata} to train and evaluate multi-aspect sentence embeddings. Further, we use AspectCSE to train multi- and single-aspect sentence embedding models for the \textit{country} and \textit{industry} aspects. For fine-tuning, we use $\textrm{BERT}_{\small{\texttt{base}}}$ and the training configuration presented in Table \ref{tab:training_configuration_simcse}. To evaluate the performance of generic sentence embeddings on the Wikipedia + Wikidata test dataset, we use a trained $\textrm{SimCSE}_{\small{\texttt{sup-bert-base}}}$ model \cite{gao-etal-2021-simcse}, which generates state-of-the-art generic sentence embeddings.

\begin{table*}[ht!]
    \centering

    \renewcommand{\arraystretch}{1.2} %
    
    \begin{tabular}{cl ccc ccc}
    
    \toprule
    \multicolumn{2}{l}{\textbf{Aspects $\rightarrow$}} & \multicolumn{3}{c}{\textit{Country}} & \multicolumn{3}{c}{\textit{Industry}} \\
    
    \cmidrule(lr){3-5}
    \cmidrule(lr){6-8}

    \multicolumn{2}{l}{\textbf{Embedding type $\downarrow$}} &      \textbf{P} &      \textbf{R} &    \textbf{MRR} &        \textbf{P} &      \textbf{R} &    \textbf{MRR}  \\
    
    \midrule
    \multicolumn{2}{l}{$\textrm{SimCSE}_{\small{\texttt{generic}}}$} & 
    0.315 & 0.058 &  0.523 &  0.320 &  0.061 &   0.531 \\
    \multicolumn{2}{l}{$\textrm{AspectCSE}_{\small{\texttt{single-aspect}}}$} & 0.390 & 0.124 & 0.558 & \textbf{0.625} & \textbf{0.178} & 0.729 \\ 
    \multicolumn{2}{l}{$\textrm{AspectCSE}_{\small{\texttt{multi-aspect(Intersection)}}}$} & 0.444&0.102&0.593 & 0.622& 0.174&0.720\\
    \multicolumn{2}{l}{$\textrm{AspectCSE}_{\small{\texttt{multi-aspect(Union)}}}$}& \textbf{0.555}&\textbf{0.163}&\textbf{0.738} &0.538&0.155&\textbf{0.747}\\
    \bottomrule
    \end{tabular}
    \caption{Evaluation results for retrieving the $k=10$ most similar elements for different sentence embedding approaches on the Wikipedia + Wikidata test dataset. Precision@k (P), Recall@k (R), and Mean Reciprocal Rank@k (MRR) are reported.} 
    \label{tab:wiki_evaluation}
\end{table*}

\begin{figure*}[hb!]
    \centering
    \includegraphics[width=\textwidth]{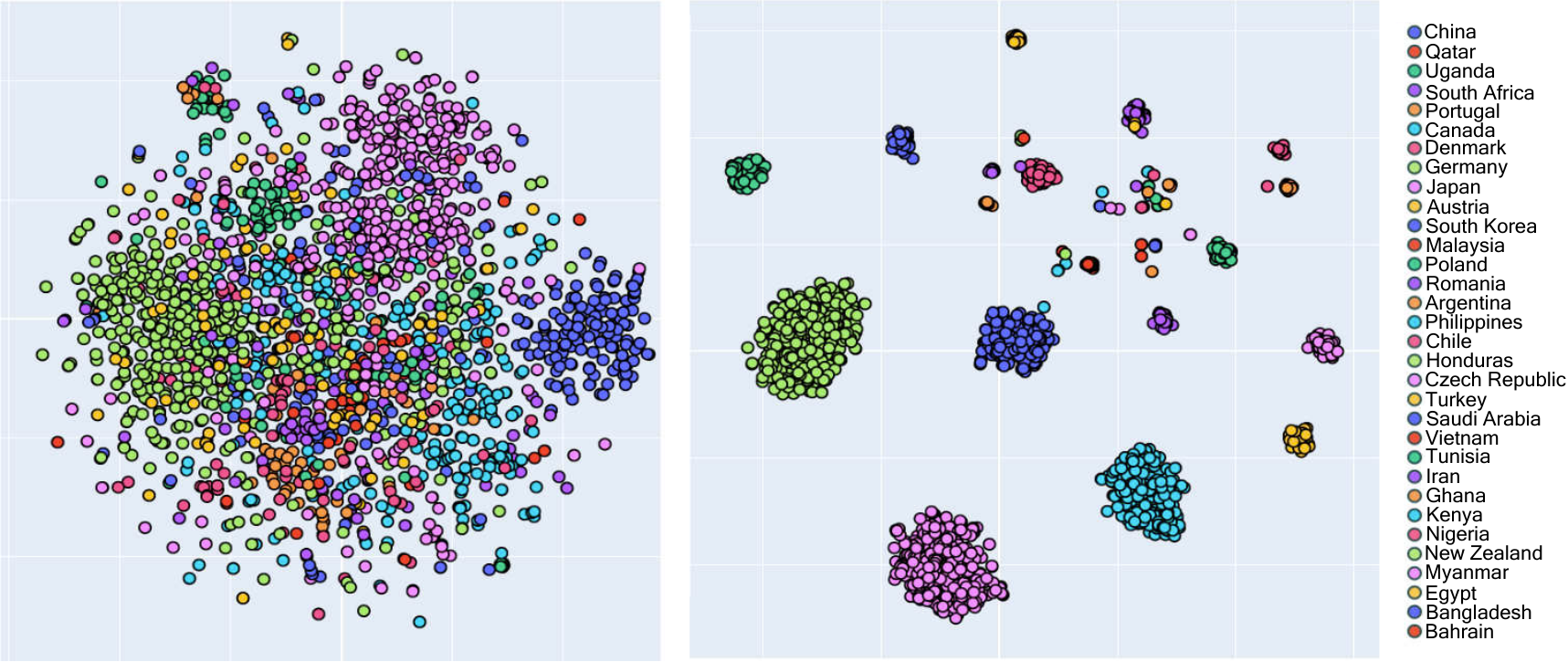}
    \caption{Comparison of generic sentence embeddings (left) vs. single-aspect sentence embeddings based on the \textit{country} aspect (right).}
    \label{fig:country_specific_embedding}
\end{figure*}

\begin{figure*}[ht!]
    \centering
    \includegraphics[width=\textwidth]{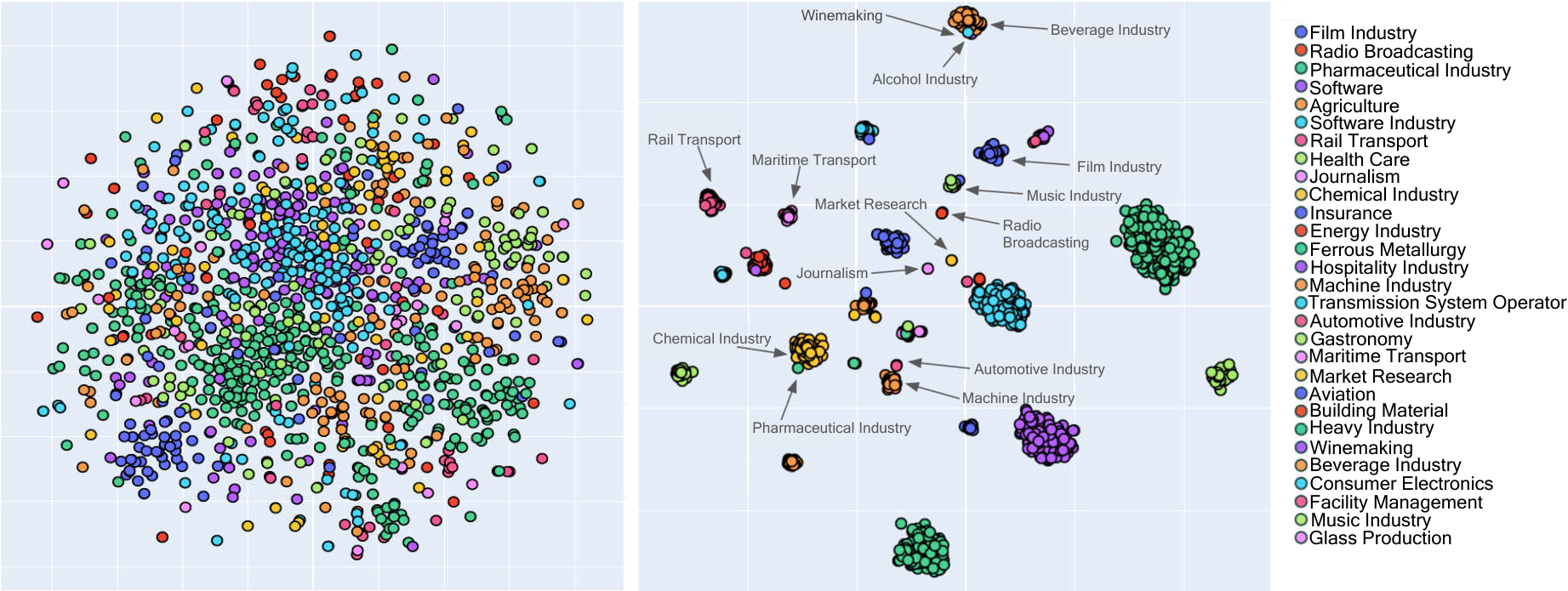}
    \caption{Comparison of generic sentence embeddings (left) vs. single-aspect sentence embeddings based on the \textit{industry} aspect (right).}
    \label{fig:industry_specific_embedding}
\end{figure*}

\section{Evaluation}

\subsection{Information Retrieval Performance}

For evaluation, we follow the approach of \citet{10.1145/3529372.3530912} and frame it as an information retrieval task. Therefore, we retrieve the $k=10$ nearest neighbors for each element in the respective test datasets. After that, we determine the number of retrieved elements that match the particular aspect label of the seed element. We use the following evaluation metrics for this purpose:

\begin{itemize}
    \item \textbf{Precision@k} (\acs{p}): The number of nearest neighbors (within the top $k$ candidates) that share the same aspect as the seed document divided by $k$.
    \item \textbf{Recall@k} (\acs{r}): The number of nearest neighbors (within the top $k$ candidates) that share the same aspect as the seed document divided by the number of labeled documents with the seed document's aspect.
    \item \textbf{Mean Reciprocal Rank@k} (\acs{mrr}): Measure of the ranking quality for the nearest neighbors, calculated by averaging the reciprocal ranks ($\frac{1}{\textrm{rank}}$) of each neighbor. This adds more weight to correctly labeled neighbors the higher they rank.
\end{itemize}

\paragraph{Papers with Code}

Table \ref{tab:pwc-evaluation} compares AspectCSE, generic sentence embedding baselines, and the aspect-based multiple negative ranking baseline. The generic sentence embedding models perform badly for all evaluated aspects. Except for SPECTER, which achieves a respectable MRR score in the \textit{dataset} aspect, generic models always perform significantly worse than aspect-based models. Therefore, aspect-based models retrieve similar texts of the same aspect much better than generic ones. Furthermore, By a large margin, AspectCSE outperforms the multiple negative ranking approach on all aspects and metrics. The average improvement is 3.97\% for MRR scores of all \ac{pwc} aspects. Hence, AspectCSE is a better approach for training aspect-based sentence embedding models. Accordingly, we use AspectCSE to train and evaluate multi-aspect sentence embedding models on the Wikipedia + Wikidata dataset. 

\paragraph{Wikipedia and Wikidata}
Table \ref{tab:wiki_evaluation} shows the evaluation results for the multi-aspect sentence embeddings on the Wikipedia + Wikidata test dataset. All AspectCSE models achieve strong performance in both aspects. While we train two separate embedding models for the single-aspect case (one embedding model each for the \textit{country} and \textit{industry} aspects), the multi-aspect models are trained on both aspects simultaneously. Therefore, in the multi-aspect cases, only one model is used to retrieve the most similar elements for both aspects. Surprisingly, the best MRR scores for the \textit{country} and \textit{industry} aspects are achieved using the multi-aspect (Union) model, outperforming the multi-aspect (Intersection) and even the single-aspect models. A possible reason is that training sentence embedding models for multiple aspects provides the model with more training data. For example, a correlation exists between the type of industry and certain countries (e.g., Arab countries that have a higher than average density of oil companies) that may function as additional training data for the model.

\subsection{Embedding Space Exploration}

In addition to the information retrieval evaluation, we visually analyze selected generic, single-, and multi-aspect sentence embeddings. Therefore, we again use the Wikipedia + Wikidata dataset and the trained models described in Section \ref{sec:multi-aspect-sentence-embeddings}. We utilize t-SNE \cite{JMLR:v9:vandermaaten08a} to reduce the dimensionality of sentence embeddings from 768 to 2 and color all data points according to their aspect labels. Figures \ref{fig:country_specific_embedding} and \ref{fig:industry_specific_embedding} show the embedding spaces of generic and single-aspect sentence embeddings for the \textit{country} and \textit{industry} aspects. In these figures, generic sentence embeddings weakly capture both target aspects, as certain aspect labels dominate some regions. However, no clear separation can be observed between aspect labels and many aspect labels are scattered throughout the entire embedding space. Meanwhile, a sharp separation exists between aspect labels for aspect-based sentence embeddings with dense clusters of elements that share the same aspect label. This finding is consistent with our results in Table \ref{tab:wiki_evaluation}. Figure \ref{fig:industry_specific_embedding} shows the local neighborhoods of industry-specific sentence embeddings that reflect the semantic similarity of different industries. We observe that embeddings of the same aspect label are close to each other, and those of semantically similar aspect labels are closer when compared to embeddings with semantically dissimilar aspect labels. For example, embeddings with the semantically related aspect labels "Film Industry", "Music Industry", and "Radio Broadcasting" are close to each other, whereas "Rail Transport" and "Maritime Transport" are located next to each other. 

\begin{figure}[hb!]
    \centering
    \includegraphics[width=\linewidth]{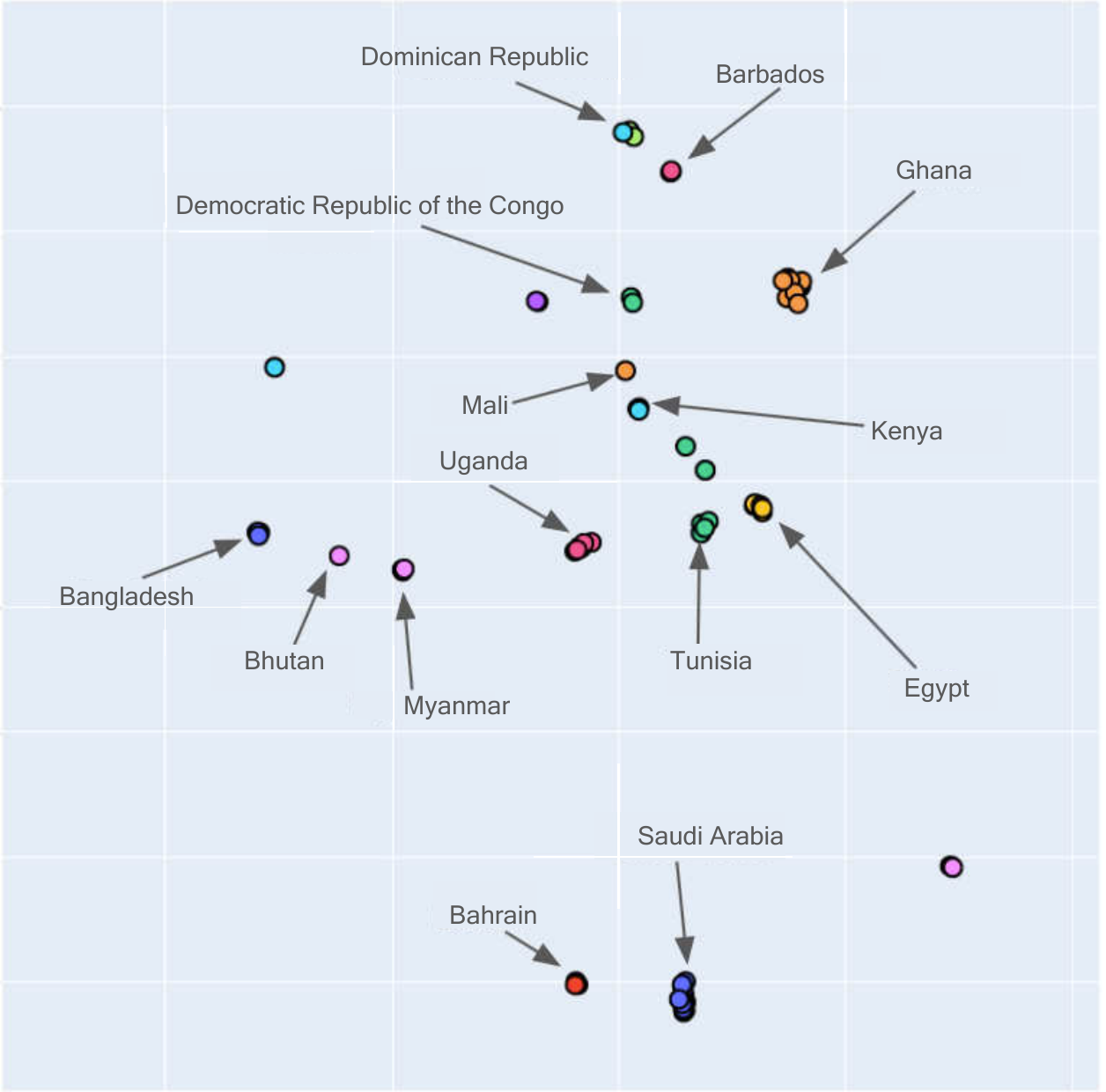}
    \caption{Local embedding space for single-aspect sentence embeddings based on the \textit{country} aspect. The colors represent different aspect labels for the \textit{country} aspect.}
    \label{fig:example_country}
\end{figure}

Figure \ref{fig:example_country} shows the local neighborhoods of single-aspect sentence embeddings based on the \textit{country} aspect. We observe a similar behavior as in Figure \ref{fig:industry_specific_embedding}, where embeddings of semantically similar aspect labels are close. For example, country-specific sentence embeddings of African countries (e.g., Kenya, Egypt, and Mali), Arab countries (e.g., Saudi Arabia, Bahrain), and South American countries (e.g., Dominican Republic, Barbados) share local neighborhoods, respectively. Although a correlation exists between semantically similar aspect labels and local neighborhoods in many cases, this pattern is not consistent for all aspect labels. For example, embeddings for the aspect label "Austria" are closer to the embeddings from "Japan" than to those for "Germany". This similarity pattern is likely a result of the fact that some texts from our training dataset are annotated with multiple aspects (e.g., "Amazon" is annotated with "e-commerce", "retail", and "cloud computing"). Since the model optimizes the embedding for Amazon to be close to e-commerce, retail, and cloud computing companies, all embeddings from these industries are pulled closer together. As the same company often operates in related industries (e.g., e-commerce and retail), this is likely why sentence embeddings of related aspect labels are close to each other. The pattern inconsistency may be partially a consequence of dimensionality reduction, where fine-grained differences between embeddings become lost.

\begin{figure}[ht!]
    \centering
    \includegraphics[width=\linewidth]{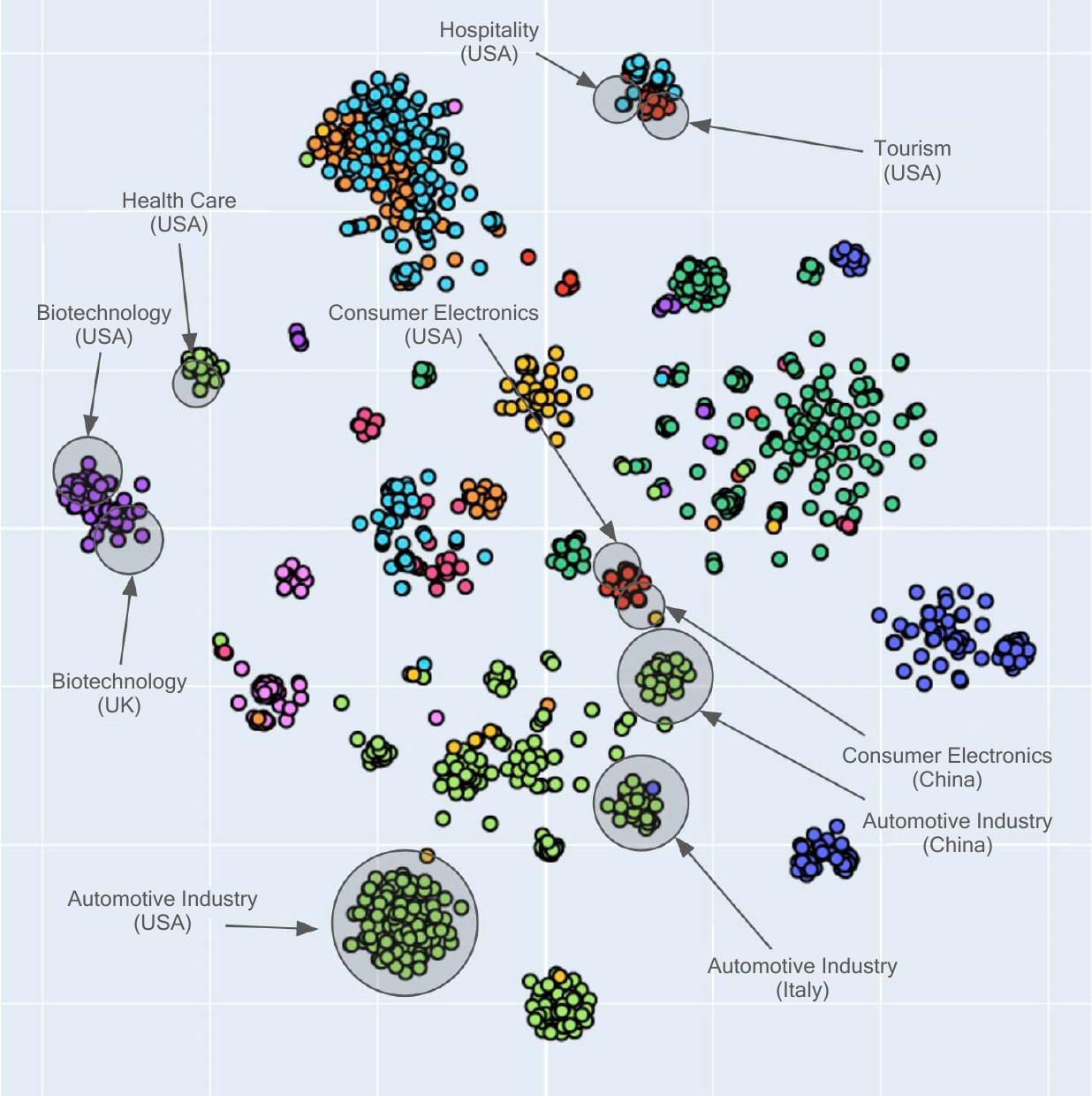}
    \caption{Global embedding space for multi-aspect sentence embeddings (Union). The colors represent different aspect labels for the \textit{industry} aspect. The aspect-based sentence embedding model is trained with the contrastive learning approach stated in Section \ref{sec:contrastive-learning} and on the Wikipedia + Wikidata dataset described in Section \ref{sec:wikipedia-wikidata}}
    \label{fig:example_mixed}
\end{figure}

Figure \ref{fig:example_mixed} shows the embeddings space for multi-aspect sentence embeddings (Union). This multi-aspect sentence embedding model (Union) learned to keep embeddings close to each other that share either the same \textit{industry} or the same \textit{country} or both aspects. As shown in the figure, only the \textit{industry} aspect is colored, as it is the more dominant aspect for the spatial positioning of embeddings. 
Figure \ref{fig:example_mixed} shows the local neighborhoods that mostly contain embeddings of the same \textit{industry} aspects. Simultaneously, the \textit{country} aspect determines the spatial positioning of embeddings within the individual "industry clusters". Sentence embeddings that belong to a certain \textit{industry} aspect, such as "Automotive" are split into different country-specific sub-clusters. Furthermore, embeddings at the boundary between industries are likely to share the same \textit{country} aspect. This is shown, for example, in "Automotive Industry (China)" and "Consumer Electronics (China)" embeddings located next to each other. 

Overall, training AspectCSE using \ac{kg} properties as aspects performs well in all our evaluations. Moreover, the multi-aspect (Union) model outperforms all other models by a large margin. Therefore, using \ac{kg} properties and AspectCSE to train single-aspect and especially multi-aspect sentence embedding models achieves meaningful results in \ac{sts} tasks. 

\section{Conclusion}

In this work, we proposed using Wikidata knowledge graph properties to train single-aspect and multi-aspect sentence embedding models. Unlike single-aspect sentence embeddings, multi-aspect sentence embeddings consider multiple specific aspects simultaneously during similarity comparisons. We regarded \ac{sts} as an information retrieval task and introduced the AspectCSE approach for training aspect-based sentence embedding models that achieve state-of-the-art performance on the \ac{pwc} benchmark dataset. Furthermore, we demonstrated that training aspect-based sentence embedding models on multiple aspects simultaneously even surpasses the performance of single-aspect sentence embedding models. Finally, we show that the semantic similarity between different aspect labels is often connected to spatial proximity in the embedding space. This behavior is even clear if we train sentence embedding models only for similarity within the same aspect label but not explicitly for similarity between different aspect labels.

\section{Limitations}

AspectCSE only works for domains and languages with pretrained language models available for fine-tuning. Furthermore, using Wikidata \ac{kg} properties to train single-aspect and multi-aspect sentence embedding models requires the availability of this structured information in large quantities. For widely used languages and domains, this requirement may be given. However, for underrepresented languages and domains, Wikidata information is sparse, which has a negative impact on AspectCSE. Moreover, we evaluated our approach using texts that comprise entire paragraphs. Whether AspectCSE can also properly represent the specific aspects contained in individual sentences needs to be investigated in future work. Finally, training AspectCSE using CPU only is not feasible. Therefore, we used a Nvidia v100 GPU for AspectCSE training.

\bibliographystyle{acl_natbib}
\bibliography{anthology,custom}

\begin{thebibliography}{46}
\expandafter\ifx\csname natexlab\endcsname\relax\def\natexlab#1{#1}\fi

\bibitem[{Agirre et~al.(2015)Agirre, Banea, Cardie, Cer, Diab, Gonzalez-Agirre,
  Guo, Lopez-Gazpio, Maritxalar, Mihalcea, Rigau, Uria, and
  Wiebe}]{agirre-etal-2015-semeval}
Eneko Agirre, Carmen Banea, Claire Cardie, Daniel Cer, Mona Diab, Aitor
  Gonzalez-Agirre, Weiwei Guo, I{\~n}igo Lopez-Gazpio, Montse Maritxalar, Rada
  Mihalcea, German Rigau, Larraitz Uria, and Janyce Wiebe. 2015.
\newblock \href {https://doi.org/10.18653/v1/S15-2045} {{S}em{E}val-2015 task
  2: Semantic textual similarity, {E}nglish, {S}panish and pilot on
  interpretability}.
\newblock In \emph{Proceedings of the 9th International Workshop on Semantic
  Evaluation ({S}em{E}val 2015)}, pages 252--263, Denver, Colorado. Association
  for Computational Linguistics.

\bibitem[{Agirre et~al.(2014)Agirre, Banea, Cardie, Cer, Diab, Gonzalez-Agirre,
  Guo, Mihalcea, Rigau, and Wiebe}]{agirre-etal-2014-semeval}
Eneko Agirre, Carmen Banea, Claire Cardie, Daniel Cer, Mona Diab, Aitor
  Gonzalez-Agirre, Weiwei Guo, Rada Mihalcea, German Rigau, and Janyce Wiebe.
  2014.
\newblock \href {https://doi.org/10.3115/v1/S14-2010} {{S}em{E}val-2014 task
  10: Multilingual semantic textual similarity}.
\newblock In \emph{Proceedings of the 8th International Workshop on Semantic
  Evaluation ({S}em{E}val 2014)}, pages 81--91, Dublin, Ireland. Association
  for Computational Linguistics.

\bibitem[{Agirre et~al.(2016)Agirre, Banea, Cer, Diab, Gonzalez-Agirre,
  Mihalcea, Rigau, and Wiebe}]{agirre-etal-2016-semeval}
Eneko Agirre, Carmen Banea, Daniel Cer, Mona Diab, Aitor Gonzalez-Agirre, Rada
  Mihalcea, German Rigau, and Janyce Wiebe. 2016.
\newblock \href {https://doi.org/10.18653/v1/S16-1081} {{S}em{E}val-2016 task
  1: Semantic textual similarity, monolingual and cross-lingual evaluation}.
\newblock In \emph{Proceedings of the 10th International Workshop on Semantic
  Evaluation ({S}em{E}val-2016)}, pages 497--511, San Diego, California.
  Association for Computational Linguistics.

\bibitem[{Agirre et~al.(2012)Agirre, Cer, Diab, and
  Gonzalez-Agirre}]{agirre-etal-2012-semeval}
Eneko Agirre, Daniel Cer, Mona Diab, and Aitor Gonzalez-Agirre. 2012.
\newblock \href {https://aclanthology.org/S12-1051} {{S}em{E}val-2012 task 6: A
  pilot on semantic textual similarity}.
\newblock In \emph{*{SEM} 2012: The First Joint Conference on Lexical and
  Computational Semantics {--} Volume 1: Proceedings of the main conference and
  the shared task, and Volume 2: Proceedings of the Sixth International
  Workshop on Semantic Evaluation ({S}em{E}val 2012)}, pages 385--393,
  Montr{\'e}al, Canada. Association for Computational Linguistics.

\bibitem[{Agirre et~al.(2013)Agirre, Cer, Diab, Gonzalez-Agirre, and
  Guo}]{agirre-etal-2013-sem}
Eneko Agirre, Daniel Cer, Mona Diab, Aitor Gonzalez-Agirre, and Weiwei Guo.
  2013.
\newblock \href {https://aclanthology.org/S13-1004} {*{SEM} 2013 shared task:
  Semantic textual similarity}.
\newblock In \emph{Second Joint Conference on Lexical and Computational
  Semantics (*{SEM}), Volume 1: Proceedings of the Main Conference and the
  Shared Task: Semantic Textual Similarity}, pages 32--43, Atlanta, Georgia,
  USA. Association for Computational Linguistics.

\bibitem[{B{\"a}r et~al.(2011)B{\"a}r, Zesch, and
  Gurevych}]{bar-etal-2011-reflective}
Daniel B{\"a}r, Torsten Zesch, and Iryna Gurevych. 2011.
\newblock \href {https://aclanthology.org/R11-1071} {A reflective view on text
  similarity}.
\newblock In \emph{Proceedings of the International Conference Recent Advances
  in Natural Language Processing 2011}, pages 515--520, Hissar, Bulgaria.
  Association for Computational Linguistics.

\bibitem[{Beltagy et~al.(2019)Beltagy, Lo, and
  Cohan}]{beltagy-etal-2019-scibert}
Iz~Beltagy, Kyle Lo, and Arman Cohan. 2019.
\newblock \href {https://doi.org/10.18653/v1/D19-1371} {{S}ci{BERT}: A
  pretrained language model for scientific text}.
\newblock In \emph{Proceedings of the 2019 Conference on Empirical Methods in
  Natural Language Processing and the 9th International Joint Conference on
  Natural Language Processing (EMNLP-IJCNLP)}, pages 3615--3620, Hong Kong,
  China. Association for Computational Linguistics.

\bibitem[{Brun and Nikoulina(2018)}]{brun-nikoulina-2018-aspect}
Caroline Brun and Vassilina Nikoulina. 2018.
\newblock \href {https://doi.org/10.18653/v1/W18-6217} {Aspect based sentiment
  analysis into the wild}.
\newblock In \emph{Proceedings of the 9th Workshop on Computational Approaches
  to Subjectivity, Sentiment and Social Media Analysis}, pages 116--122,
  Brussels, Belgium. Association for Computational Linguistics.

\bibitem[{Cer et~al.(2017)Cer, Diab, Agirre, Lopez-Gazpio, and
  Specia}]{cer-etal-2017-semeval}
Daniel Cer, Mona Diab, Eneko Agirre, I{\~n}igo Lopez-Gazpio, and Lucia Specia.
  2017.
\newblock \href {https://doi.org/10.18653/v1/S17-2001} {{S}em{E}val-2017 task
  1: Semantic textual similarity multilingual and crosslingual focused
  evaluation}.
\newblock In \emph{Proceedings of the 11th International Workshop on Semantic
  Evaluation ({S}em{E}val-2017)}, pages 1--14, Vancouver, Canada. Association
  for Computational Linguistics.

\bibitem[{Cer et~al.(2018)Cer, Yang, Kong, Hua, Limtiaco, St.~John, Constant,
  Guajardo-Cespedes, Yuan, Tar, Strope, and Kurzweil}]{cer-etal-2018-universal}
Daniel Cer, Yinfei Yang, Sheng-yi Kong, Nan Hua, Nicole Limtiaco, Rhomni
  St.~John, Noah Constant, Mario Guajardo-Cespedes, Steve Yuan, Chris Tar,
  Brian Strope, and Ray Kurzweil. 2018.
\newblock \href {https://doi.org/10.18653/v1/D18-2029} {Universal sentence
  encoder for {E}nglish}.
\newblock In \emph{Proceedings of the 2018 Conference on Empirical Methods in
  Natural Language Processing: System Demonstrations}, pages 169--174,
  Brussels, Belgium. Association for Computational Linguistics.

\bibitem[{Chan et~al.(2018)Chan, Chang, Hope, Shahaf, and
  Kittur}]{10.1145/3274300}
Joel Chan, Joseph~Chee Chang, Tom Hope, Dafna Shahaf, and Aniket Kittur. 2018.
\newblock \href {https://doi.org/10.1145/3274300} {Solvent: A mixed initiative
  system for finding analogies between research papers}.
\newblock \emph{Proc. ACM Hum.-Comput. Interact.}, 2(CSCW).

\bibitem[{Chuang et~al.(2022)Chuang, Dangovski, Luo, Zhang, Chang, Soljacic,
  Li, Yih, Kim, and Glass}]{chuang-etal-2022-diffcse}
Yung-Sung Chuang, Rumen Dangovski, Hongyin Luo, Yang Zhang, Shiyu Chang, Marin
  Soljacic, Shang-Wen Li, Scott Yih, Yoon Kim, and James Glass. 2022.
\newblock \href {https://doi.org/10.18653/v1/2022.naacl-main.311} {{D}iff{CSE}:
  Difference-based contrastive learning for sentence embeddings}.
\newblock In \emph{Proceedings of the 2022 Conference of the North American
  Chapter of the Association for Computational Linguistics: Human Language
  Technologies}, pages 4207--4218, Seattle, United States. Association for
  Computational Linguistics.

\bibitem[{Cohan et~al.(2020)Cohan, Feldman, Beltagy, Downey, and
  Weld}]{cohan-etal-2020-specter}
Arman Cohan, Sergey Feldman, Iz~Beltagy, Doug Downey, and Daniel Weld. 2020.
\newblock \href {https://doi.org/10.18653/v1/2020.acl-main.207} {{SPECTER}:
  Document-level representation learning using citation-informed transformers}.
\newblock In \emph{Proceedings of the 58th Annual Meeting of the Association
  for Computational Linguistics}, pages 2270--2282, Online. Association for
  Computational Linguistics.

\bibitem[{Conneau et~al.(2017)Conneau, Kiela, Schwenk, Barrault, and
  Bordes}]{conneau-etal-2017-supervised}
Alexis Conneau, Douwe Kiela, Holger Schwenk, Lo{\"\i}c Barrault, and Antoine
  Bordes. 2017.
\newblock \href {https://doi.org/10.18653/v1/D17-1070} {Supervised learning of
  universal sentence representations from natural language inference data}.
\newblock In \emph{Proceedings of the 2017 Conference on Empirical Methods in
  Natural Language Processing}, pages 670--680, Copenhagen, Denmark.
  Association for Computational Linguistics.

\bibitem[{Devlin et~al.(2019)Devlin, Chang, Lee, and
  Toutanova}]{devlin-etal-2019-bert}
Jacob Devlin, Ming-Wei Chang, Kenton Lee, and Kristina Toutanova. 2019.
\newblock \href {https://doi.org/10.18653/v1/N19-1423} {{BERT}: Pre-training of
  deep bidirectional transformers for language understanding}.
\newblock In \emph{Proceedings of the 2019 Conference of the North {A}merican
  Chapter of the Association for Computational Linguistics: Human Language
  Technologies, Volume 1 (Long and Short Papers)}, pages 4171--4186,
  Minneapolis, Minnesota. Association for Computational Linguistics.

\bibitem[{Gao et~al.(2021)Gao, Yao, and Chen}]{gao-etal-2021-simcse}
Tianyu Gao, Xingcheng Yao, and Danqi Chen. 2021.
\newblock \href {https://doi.org/10.18653/v1/2021.emnlp-main.552} {{S}im{CSE}:
  Simple contrastive learning of sentence embeddings}.
\newblock In \emph{Proceedings of the 2021 Conference on Empirical Methods in
  Natural Language Processing}, pages 6894--6910, Online and Punta Cana,
  Dominican Republic. Association for Computational Linguistics.

\bibitem[{Giorgi et~al.(2021)Giorgi, Nitski, Wang, and
  Bader}]{giorgi-etal-2021-declutr}
John Giorgi, Osvald Nitski, Bo~Wang, and Gary Bader. 2021.
\newblock \href {https://doi.org/10.18653/v1/2021.acl-long.72} {{D}e{CLUTR}:
  Deep contrastive learning for unsupervised textual representations}.
\newblock In \emph{Proceedings of the 59th Annual Meeting of the Association
  for Computational Linguistics and the 11th International Joint Conference on
  Natural Language Processing (Volume 1: Long Papers)}, pages 879--895, Online.
  Association for Computational Linguistics.

\bibitem[{Gong et~al.(2020)Gong, Shen, Yu, Chen, and
  Yu}]{gong-etal-2020-recurrent}
Hongyu Gong, Yelong Shen, Dian Yu, Jianshu Chen, and Dong Yu. 2020.
\newblock \href {https://doi.org/10.18653/v1/2020.acl-main.603} {Recurrent
  chunking mechanisms for long-text machine reading comprehension}.
\newblock In \emph{Proceedings of the 58th Annual Meeting of the Association
  for Computational Linguistics}, pages 6751--6761, Online. Association for
  Computational Linguistics.

\bibitem[{Hadsell et~al.(2006)Hadsell, Chopra, and LeCun}]{1640964}
R.~Hadsell, S.~Chopra, and Y.~LeCun. 2006.
\newblock \href {https://doi.org/10.1109/CVPR.2006.100} {Dimensionality
  reduction by learning an invariant mapping}.
\newblock In \emph{2006 IEEE Computer Society Conference on Computer Vision and
  Pattern Recognition (CVPR'06)}, volume~2, pages 1735--1742.

\bibitem[{Henderson et~al.(2017)Henderson, Al-Rfou, Strope, Sung, Lukacs, Guo,
  Kumar, Miklos, and Kurzweil}]{https://doi.org/10.48550/arxiv.1705.00652}
Matthew Henderson, Rami Al-Rfou, Brian Strope, Yun-hsuan Sung, Laszlo Lukacs,
  Ruiqi Guo, Sanjiv Kumar, Balint Miklos, and Ray Kurzweil. 2017.
\newblock \href {https://doi.org/10.48550/ARXIV.1705.00652} {Efficient natural
  language response suggestion for smart reply}.

\bibitem[{Hill et~al.(2016)Hill, Cho, and Korhonen}]{hill-etal-2016-learning}
Felix Hill, Kyunghyun Cho, and Anna Korhonen. 2016.
\newblock \href {https://doi.org/10.18653/v1/N16-1162} {Learning distributed
  representations of sentences from unlabelled data}.
\newblock In \emph{Proceedings of the 2016 Conference of the North {A}merican
  Chapter of the Association for Computational Linguistics: Human Language
  Technologies}, pages 1367--1377, San Diego, California. Association for
  Computational Linguistics.

\bibitem[{Huang et~al.(2020)Huang, Huang, Ding, Hsu, and
  Giles}]{huang-etal-2020-coda}
Ting-Hao~Kenneth Huang, Chieh-Yang Huang, Chien-Kuang~Cornelia Ding, Yen-Chia
  Hsu, and C.~Lee Giles. 2020.
\newblock \href {https://aclanthology.org/2020.nlpcovid19-acl.6} {{CODA-19}:
  Using a non-expert crowd to annotate research aspects on 10,000+ abstracts in
  the {COVID-19} open research dataset}.
\newblock In \emph{Proceedings of the 1st Workshop on {NLP} for {COVID-19} at
  {ACL} 2020}, Online. Association for Computational Linguistics.

\bibitem[{Kiros et~al.(2015)Kiros, Zhu, Salakhutdinov, Zemel, Urtasun,
  Torralba, and Fidler}]{NIPS2015_f442d33f}
Ryan Kiros, Yukun Zhu, Russ~R Salakhutdinov, Richard Zemel, Raquel Urtasun,
  Antonio Torralba, and Sanja Fidler. 2015.
\newblock \href
  {https://proceedings.neurips.cc/paper/2015/file/f442d33fa06832082290ad8544a8da27-Paper.pdf}
  {Skip-thought vectors}.
\newblock In \emph{Advances in Neural Information Processing Systems},
  volume~28. Curran Associates, Inc.

\bibitem[{Kobayashi et~al.(2018)Kobayashi, Shimbo, and
  Matsumoto}]{10.1145/3197026.3197059}
Yuta Kobayashi, Masashi Shimbo, and Yuji Matsumoto. 2018.
\newblock \href {https://doi.org/10.1145/3197026.3197059} {Citation
  recommendation using distributed representation of discourse facets in
  scientific articles}.
\newblock In \emph{Proceedings of the 18th ACM/IEEE on Joint Conference on
  Digital Libraries}, JCDL '18, page 243–251, New York, NY, USA. Association
  for Computing Machinery.

\bibitem[{Liang et~al.(2022)Liang, Wei, Mao, Wang, and
  He}]{liang-etal-2022-bisyn}
Shuo Liang, Wei Wei, Xian-Ling Mao, Fei Wang, and Zhiyong He. 2022.
\newblock \href {https://doi.org/10.18653/v1/2022.findings-acl.144}
  {{B}i{S}yn-{GAT}+: Bi-syntax aware graph attention network for aspect-based
  sentiment analysis}.
\newblock In \emph{Findings of the Association for Computational Linguistics:
  ACL 2022}, pages 1835--1848, Dublin, Ireland. Association for Computational
  Linguistics.

\bibitem[{Logeswaran and Lee(2018)}]{logeswaran2018an}
Lajanugen Logeswaran and Honglak Lee. 2018.
\newblock \href {https://openreview.net/forum?id=rJvJXZb0W} {An efficient
  framework for learning sentence representations}.
\newblock In \emph{International Conference on Learning Representations}.

\bibitem[{van~der Maaten and Hinton(2008)}]{JMLR:v9:vandermaaten08a}
Laurens van~der Maaten and Geoffrey Hinton. 2008.
\newblock \href {http://jmlr.org/papers/v9/vandermaaten08a.html} {Visualizing
  data using t-sne}.
\newblock \emph{Journal of Machine Learning Research}, 9(86):2579--2605.

\bibitem[{Marelli et~al.(2014)Marelli, Menini, Baroni, Bentivogli, Bernardi,
  and Zamparelli}]{marelli-etal-2014-sick}
Marco Marelli, Stefano Menini, Marco Baroni, Luisa Bentivogli, Raffaella
  Bernardi, and Roberto Zamparelli. 2014.
\newblock \href
  {http://www.lrec-conf.org/proceedings/lrec2014/pdf/363_Paper.pdf} {A {SICK}
  cure for the evaluation of compositional distributional semantic models}.
\newblock In \emph{Proceedings of the Ninth International Conference on
  Language Resources and Evaluation ({LREC}'14)}, pages 216--223, Reykjavik,
  Iceland. European Language Resources Association (ELRA).

\bibitem[{Ostendorff et~al.(2022)Ostendorff, Blume, Ruas, Gipp, and
  Rehm}]{10.1145/3529372.3530912}
Malte Ostendorff, Till Blume, Terry Ruas, Bela Gipp, and Georg Rehm. 2022.
\newblock \href {https://doi.org/10.1145/3529372.3530912} {Specialized document
  embeddings for aspect-based similarity of research papers}.
\newblock In \emph{Proceedings of the 22nd ACM/IEEE Joint Conference on Digital
  Libraries}, JCDL '22, New York, NY, USA. Association for Computing Machinery.

\bibitem[{Ostendorff et~al.(2020{\natexlab{a}})Ostendorff, Ruas, Blume, Gipp,
  and Rehm}]{ostendorff-etal-2020-aspect}
Malte Ostendorff, Terry Ruas, Till Blume, Bela Gipp, and Georg Rehm.
  2020{\natexlab{a}}.
\newblock \href {https://doi.org/10.18653/v1/2020.coling-main.545}
  {Aspect-based document similarity for research papers}.
\newblock In \emph{Proceedings of the 28th International Conference on
  Computational Linguistics}, pages 6194--6206, Barcelona, Spain (Online).
  International Committee on Computational Linguistics.

\bibitem[{Ostendorff et~al.(2020{\natexlab{b}})Ostendorff, Ruas, Schubotz,
  Rehm, and Gipp}]{10.1145/3383583.3398525}
Malte Ostendorff, Terry Ruas, Moritz Schubotz, Georg Rehm, and Bela Gipp.
  2020{\natexlab{b}}.
\newblock \href {https://doi.org/10.1145/3383583.3398525} {Pairwise multi-class
  document classification for semantic relations between wikipedia articles}.
\newblock In \emph{Proceedings of the ACM/IEEE Joint Conference on Digital
  Libraries in 2020}, JCDL '20, page 127–136, New York, NY, USA. Association
  for Computing Machinery.

\bibitem[{Pontiki et~al.(2014)Pontiki, Galanis, Pavlopoulos, Papageorgiou,
  Androutsopoulos, and Manandhar}]{pontiki-etal-2014-semeval}
Maria Pontiki, Dimitris Galanis, John Pavlopoulos, Harris Papageorgiou, Ion
  Androutsopoulos, and Suresh Manandhar. 2014.
\newblock \href {https://doi.org/10.3115/v1/S14-2004} {{S}em{E}val-2014 task 4:
  Aspect based sentiment analysis}.
\newblock In \emph{Proceedings of the 8th International Workshop on Semantic
  Evaluation ({S}em{E}val 2014)}, pages 27--35, Dublin, Ireland. Association
  for Computational Linguistics.

\bibitem[{Reimers and Gurevych(2019)}]{reimers-gurevych-2019-sentence}
Nils Reimers and Iryna Gurevych. 2019.
\newblock \href {https://doi.org/10.18653/v1/D19-1410} {Sentence-{BERT}:
  Sentence embeddings using {S}iamese {BERT}-networks}.
\newblock In \emph{Proceedings of the 2019 Conference on Empirical Methods in
  Natural Language Processing and the 9th International Joint Conference on
  Natural Language Processing (EMNLP-IJCNLP)}, pages 3982--3992, Hong Kong,
  China. Association for Computational Linguistics.

\bibitem[{Reimers and Gurevych(2020)}]{reimers-gurevych-2020-making}
Nils Reimers and Iryna Gurevych. 2020.
\newblock \href {https://doi.org/10.18653/v1/2020.emnlp-main.365} {Making
  monolingual sentence embeddings multilingual using knowledge distillation}.
\newblock In \emph{Proceedings of the 2020 Conference on Empirical Methods in
  Natural Language Processing (EMNLP)}, pages 4512--4525, Online. Association
  for Computational Linguistics.

\bibitem[{Schneider et~al.(2022)Schneider, Schopf, Vladika, Galkin, Simperl,
  and Matthes}]{schneider-etal-2022-decade}
Phillip Schneider, Tim Schopf, Juraj Vladika, Mikhail Galkin, Elena Simperl,
  and Florian Matthes. 2022.
\newblock \href {https://aclanthology.org/2022.aacl-main.46} {A decade of
  knowledge graphs in natural language processing: A survey}.
\newblock In \emph{Proceedings of the 2nd Conference of the Asia-Pacific
  Chapter of the Association for Computational Linguistics and the 12th
  International Joint Conference on Natural Language Processing (Volume 1: Long
  Papers)}, pages 601--614, Online only. Association for Computational
  Linguistics.

\bibitem[{Schopf et~al.(2023{\natexlab{a}})Schopf, Arabi, and
  Matthes}]{schopf2023exploring}
Tim Schopf, Karim Arabi, and Florian Matthes. 2023{\natexlab{a}}.
\newblock \href {http://arxiv.org/abs/2307.10652} {Exploring the landscape of
  natural language processing research}.

\bibitem[{Schopf et~al.(2021)Schopf, Braun, and
  Matthes}]{schopf_etal_webist_21}
Tim Schopf, Daniel Braun, and Florian Matthes. 2021.
\newblock \href {https://doi.org/10.5220/0010710300003058} {Lbl2vec: An
  embedding-based approach for unsupervised document retrieval on predefined
  topics}.
\newblock In \emph{Proceedings of the 17th International Conference on Web
  Information Systems and Technologies - WEBIST}, pages 124--132. INSTICC,
  SciTePress.

\bibitem[{Schopf et~al.(2023{\natexlab{b}})Schopf, Braun, and
  Matthes}]{10.1145/3582768.3582795}
Tim Schopf, Daniel Braun, and Florian Matthes. 2023{\natexlab{b}}.
\newblock \href {https://doi.org/10.1145/3582768.3582795} {Evaluating
  unsupervised text classification: Zero-shot and similarity-based approaches}.
\newblock In \emph{Proceedings of the 2022 6th International Conference on
  Natural Language Processing and Information Retrieval}, NLPIR '22, page
  6–15, New York, NY, USA. Association for Computing Machinery.

\bibitem[{Schopf et~al.(2023{\natexlab{c}})Schopf, Braun, and
  Matthes}]{10.1007/978-3-031-24197-0_4}
Tim Schopf, Daniel Braun, and Florian Matthes. 2023{\natexlab{c}}.
\newblock \href {https://doi.org/https://doi.org/10.1007/978-3-031-24197-0_4}
  {Semantic label representations with lbl2vec: A similarity-based approach
  for unsupervised text classification}.
\newblock In \emph{Web Information Systems and Technologies}, pages 59--73,
  Cham. Springer International Publishing.

\bibitem[{Schopf et~al.(2022)Schopf, Klimek, and Matthes}]{schopf_etal_kdir_22}
Tim Schopf, Simon Klimek, and Florian Matthes. 2022.
\newblock \href {https://doi.org/10.5220/0011546600003335} {Patternrank:
  Leveraging pretrained language models and part of speech for unsupervised
  keyphrase extraction}.
\newblock In \emph{Proceedings of the 14th International Joint Conference on
  Knowledge Discovery, Knowledge Engineering and Knowledge Management (IC3K
  2022) - KDIR}, pages 243--248. INSTICC, SciTePress.

\bibitem[{Schopf et~al.(2023{\natexlab{d}})Schopf, Schneider, and
  Matthes}]{schopf2023efficient}
Tim Schopf, Dennis Schneider, and Florian Matthes. 2023{\natexlab{d}}.
\newblock \href {http://arxiv.org/abs/2307.03104} {Efficient domain adaptation
  of sentence embeddings using adapters}.

\bibitem[{Vrande\v{c}i\'{c} and Kr\"{o}tzsch(2014)}]{10.1145/2629489}
Denny Vrande\v{c}i\'{c} and Markus Kr\"{o}tzsch. 2014.
\newblock \href {https://doi.org/10.1145/2629489} {Wikidata: A free
  collaborative knowledgebase}.
\newblock \emph{Commun. ACM}, 57(10):78–85.

\bibitem[{Wang et~al.(2020)Wang, Lo, Chandrasekhar, Reas, Yang, Burdick, Eide,
  Funk, Katsis, Kinney, Li, Liu, Merrill, Mooney, Murdick, Rishi, Sheehan,
  Shen, Stilson, Wade, Wang, Wang, Wilhelm, Xie, Raymond, Weld, Etzioni, and
  Kohlmeier}]{wang-etal-2020-cord}
Lucy~Lu Wang, Kyle Lo, Yoganand Chandrasekhar, Russell Reas, Jiangjiang Yang,
  Doug Burdick, Darrin Eide, Kathryn Funk, Yannis Katsis, Rodney~Michael
  Kinney, Yunyao Li, Ziyang Liu, William Merrill, Paul Mooney, Dewey~A.
  Murdick, Devvret Rishi, Jerry Sheehan, Zhihong Shen, Brandon Stilson, Alex~D.
  Wade, Kuansan Wang, Nancy Xin~Ru Wang, Christopher Wilhelm, Boya Xie,
  Douglas~M. Raymond, Daniel~S. Weld, Oren Etzioni, and Sebastian Kohlmeier.
  2020.
\newblock \href {https://aclanthology.org/2020.nlpcovid19-acl.1} {{CORD-19}:
  The {COVID-19} open research dataset}.
\newblock In \emph{Proceedings of the 1st Workshop on {NLP} for {COVID-19} at
  {ACL} 2020}, Online. Association for Computational Linguistics.

\bibitem[{Xue and Li(2018)}]{xue-li-2018-aspect}
Wei Xue and Tao Li. 2018.
\newblock \href {https://doi.org/10.18653/v1/P18-1234} {Aspect based sentiment
  analysis with gated convolutional networks}.
\newblock In \emph{Proceedings of the 56th Annual Meeting of the Association
  for Computational Linguistics (Volume 1: Long Papers)}, pages 2514--2523,
  Melbourne, Australia. Association for Computational Linguistics.

\bibitem[{Yan et~al.(2021)Yan, Dai, Ji, Qiu, and Zhang}]{yan-etal-2021-unified}
Hang Yan, Junqi Dai, Tuo Ji, Xipeng Qiu, and Zheng Zhang. 2021.
\newblock \href {https://doi.org/10.18653/v1/2021.acl-long.188} {A unified
  generative framework for aspect-based sentiment analysis}.
\newblock In \emph{Proceedings of the 59th Annual Meeting of the Association
  for Computational Linguistics and the 11th International Joint Conference on
  Natural Language Processing (Volume 1: Long Papers)}, pages 2416--2429,
  Online. Association for Computational Linguistics.

\bibitem[{Zhang et~al.(2021)Zhang, Li, Deng, Bing, and
  Lam}]{zhang-etal-2021-towards-generative}
Wenxuan Zhang, Xin Li, Yang Deng, Lidong Bing, and Wai Lam. 2021.
\newblock \href {https://doi.org/10.18653/v1/2021.acl-short.64} {Towards
  generative aspect-based sentiment analysis}.
\newblock In \emph{Proceedings of the 59th Annual Meeting of the Association
  for Computational Linguistics and the 11th International Joint Conference on
  Natural Language Processing (Volume 2: Short Papers)}, pages 504--510,
  Online. Association for Computational Linguistics.

\end{thebibliography}

\end{document}